\documentclass[11pt]{article}

\usepackage[letterpaper,margin=1in,top=0.8in]{geometry}
\usepackage{newtxtext}
\usepackage{newtxmath}
\usepackage{amsmath}
\usepackage{graphicx}
\usepackage{booktabs}
\usepackage[table]{xcolor}
\usepackage{caption}
\usepackage{wrapfig}
\usepackage{natbib}
\usepackage{float}
\usepackage{hyperref}
\usepackage{caption}

\definecolor{new_yellow}{HTML}{fcb72b}
\definecolor{new_blue}{HTML}{110599}
\definecolor{new_red}{HTML}{de4d43}
\definecolor{GridColor}{HTML}{ffe6e6} 
\definecolor{RandomColor}{HTML}{e6f0ff}  

\hypersetup{
    colorlinks=true,
    linkcolor=new_red,
    citecolor=new_blue,
    urlcolor=new_yellow
}
\newcommand{\grid}[1]{\cellcolor{GridColor} #1}
\newcommand{\entire}[1]{\cellcolor{RandomColor} #1}
\newcommand{\gridOutside}[1]{\colorbox{GridColor}{#1}}
\newcommand{\entireOutside}[1]{\colorbox{RandomColor}{#1}}
\newcommand{\titleSize}{\fontsize{17.3}{20}\selectfont}

\title{Attention to Detail: Global-Local Attention for High-Resolution AI-Generated Image Detection}
\author{Lawrence Han}

\begin{document}

\makeatletter
\renewcommand{\maketitle}{
  \begin{flushleft}
    {\titleSize \textbf{Attention to Detail: Global-Local Attention for High-Resolution AI-Generated Image Detection} \par}
    \bigskip
    {\centering \large \textbf{Lawrence Han} \par}
    \smallskip
    {\centering \normalsize \texttt{lawrence.hanyl@gmail.com} \par}
    \smallskip
    {\centering \normalsize \href{https://github.com/lawrencehan5/Attention-To-Detail}{\texttt{GitHub Project Code}} \par}
    \smallskip
  \end{flushleft}
}

\renewenvironment{abstract}{
  \normalsize
  \begin{center}
    \large \textbf{Abstract}
  \end{center}
  \begin{quote}
}{
  \end{quote}
}
\makeatother

\maketitle

\begin{abstract}
\vspace{-3mm}
The rapid development of generative AI has made AI-generated images increasingly realistic and high-resolution. Most AI-generated image detection architectures typically downsample images before inputting them into models, risking the loss of fine-grained details. This paper presents \textbf{GLASS}~(\textbf{G}lobal-\textbf{L}ocal \textbf{A}ttention with \textbf{S}tratified \textbf{S}ampling), an architecture that combines a globally resized view with multiple randomly sampled local crops. These crops are original-resolution regions efficiently selected through spatially stratified sampling and aggregated using attention-based scoring. GLASS can be integrated into vision models to leverage both global and local information in images of any size. Vision Transformer, ResNet, and ConvNeXt models are used as backbones, and experiments show that GLASS outperforms standard transfer learning by achieving higher predictive performance within feasible computational constraints.
\end{abstract}

\section{Introduction}
``Is this AI?'' has increasingly become an almost reflexive response to any piece of media we see online. The rapid growth of Artificial Intelligence (AI) has enabled models to generate highly realistic, high-resolution images that are almost indistinguishable from real ones to the human eye. And with such generative AI being widely accessible, the associated risks and concerns have grown just as quickly.

\medskip
\begin{wrapfigure}{l}{0.43\textwidth}
    \vspace{-5mm}
    \includegraphics[width=0.43\textwidth]{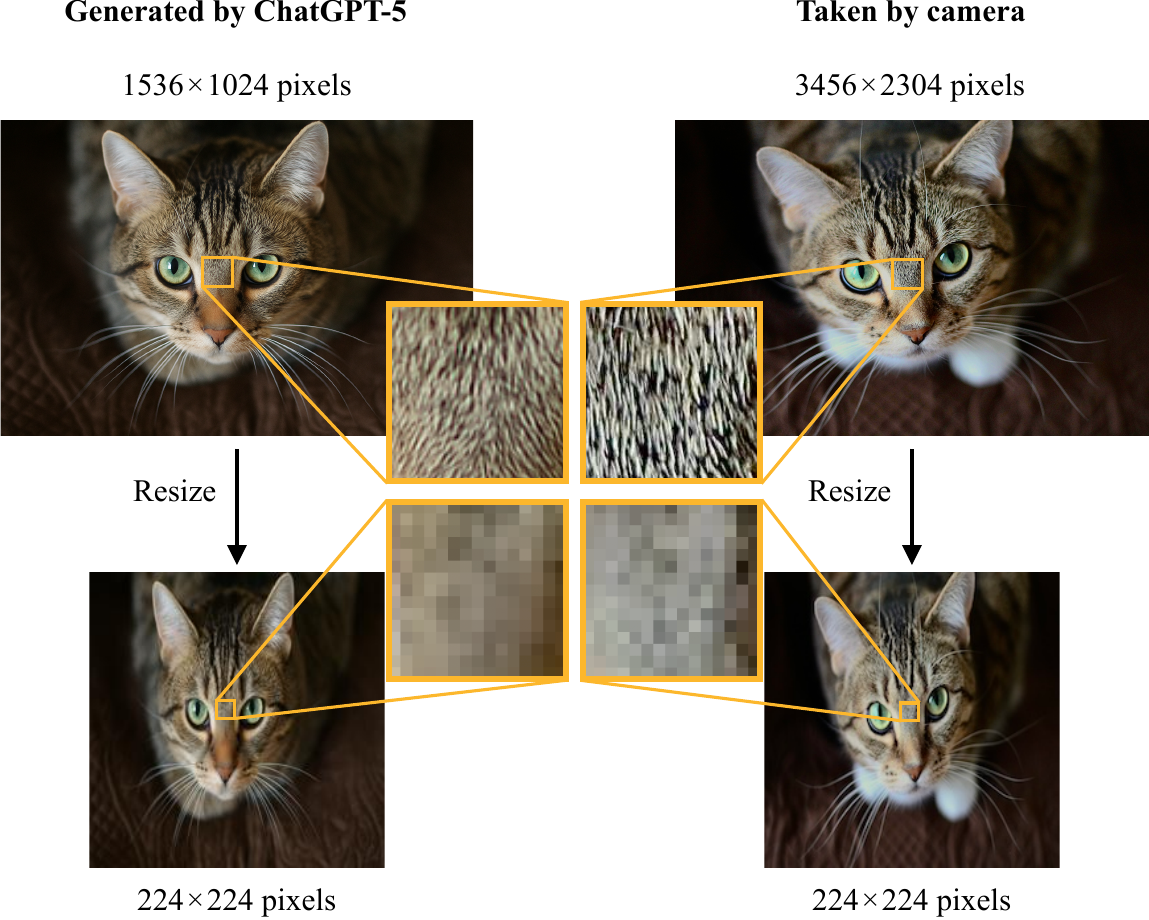}
    \vspace{-5mm}
    \caption{\textbf{Pixel-level comparison of ChatGPT-5 generated image and real image before and after resizing.} Fine details are lost after resizing.}
    \vspace{-3mm}
\end{wrapfigure}

Many architectures for AI-generated image detection typically preprocess the image by resizing it to a fixed resolution, most commonly $224\times224$ pixels, regardless of the original image size~\citep{mu2025pixelleftbehinddetailpreserving}. This ensures that all images are compatible with the base model's input requirements. However, this method can inadvertently discard fine-grained details produced by the image generator, especially in high-resolution images. Consequently, the subsequent classification model could lose relevant information that would otherwise be useful when making a more accurate decision~\citep{wang2020cnngeneratedimagessurprisinglyeasy}. 

\medskip
This paper introduces \textbf{GLASS}~(\textbf{G}lobal-\textbf{L}ocal \textbf{A}ttention with \textbf{S}tratified \textbf{S}ampling), an architecture that allows models to leverage both macro and micro information in images. A globally resized view enables models to capture semantic context, while local crops allow them to focus on fine-grained details. Local crops are randomly sampled at the original resolution, with spatial stratification applied to ensure a minimum number of crops for a large non-overlapping area. The crops are aggregated using an attention-based weighting~\citep{bahdanau2016neuralmachinetranslationjointly} mechanism to assign higher importance to regions with more informative features. This architecture is evaluated using three backbone models: Vision Transformer (ViT)~\citep{dosovitskiy2021imageworth16x16words}, ConvNeXt~\citep{liu2022convnet2020s}, and ResNet~\citep{he2015deepresiduallearningimage}. Its performance is compared to the standard transfer learning using only the global resized input.

\medskip
The contributions of this paper are: \textbf{(1)} Introduction of the GLASS architecture that utilizes stratified sampling and attention-based aggregation of sampled crops. \textbf{(2)} A comprehensive evaluation of performance comparing GLASS against standard transfer learning across three different backbone models.

\section{Related Work}
This section reviews prior work on AI-generated image detection that uses both global and local information, primarily focusing on how local crops are used and the strategies for selecting local regions.

\subsection{Applying Local Crops}
Work in AI-generated image detection has explored the use of local high-resolution crops to preserve subtle artifacts that may be lost during downsampling. The architecture introduced by~\citet{ju2022fusinggloballocalfeatures} employs a two-branch system with a multi-head attention mechanism to combine global and local features. AIDE~\citep{yan2025sanitycheckaigeneratedimage} uses ResNet-50 and CLIP-ConvNeXt as the backbone models and incorporates global-local features with Discrete Cosine Transform scoring for low-level artifact detection. Patch shuffling~\citep{chu2025semanticsregulaterethinkingpatch} randomly permutes local patches to improve generalization. Smash\&Reconstruction~\citep{zhong2024patchcraftexploringtexturepatch} randomly crops an image into local patches and classifies only on the reconstructed image of patches. In contrast, HiDA-Net~\citep{mu2025pixelleftbehinddetailpreserving} uses a ViT backbone and a feature aggregation module to combine a resized global view with original-size local crops that cover the entire image to ensure all pixels are used. GLASS also uses a two-branch system with an attention mechanism for feature aggregation, but uses additive attention scoring~\citep{bahdanau2016neuralmachinetranslationjointly, ilse2018attentionbaseddeepmultipleinstance} on the local crops only for importance weighting.

\subsection{Local Crop Selection}
While several methods focus on the use of local crops, the challenge of selecting informative crops efficiently is also important. The PSM~\citep{ju2022fusinggloballocalfeatures} module selects the most informative crops through patch scoring, rather than random selection. Its crop size is determined by a window size and then resized before being input to ResNet-50. HiDA-Net~\citep{mu2025pixelleftbehinddetailpreserving} samples the entire image at the original resolution to ensure total coverage of the local crops. TextureCrop~\citep{konstantinidou2025texturecropenhancingsyntheticimage} selects the top 10 crops using a texture selection measure. GLASS uses random selection with spatial stratification to offer a good balance between efficiency and robustness.  

\section{Methodology}

\subsection{GLASS Overview}
The GLASS architecture employs a two-stream design~\citep{simonyan2014twostreamconvolutionalnetworksaction}. The global stream processes the resized full image to capture high-level semantic content, while the local stream focuses on local crops of the original resolution to detect fine-grained detail. This two-stream design reduces the risk of a single model having to differentiate between globally resized images and locally cropped images. Two separate models, each designated and specialized for its own task, can learn weights specifically for its image scale and function. The key component of GLASS is the use of a local sampler and a simple attention mechanism. The local sampler dynamically extracts crops from the image based on the image's size. The attention mechanism focuses on features extracted from multiple local crops to aggregate them into a single highly informative one. The global image is input into the global model. Its output is concatenated with the aggregated local features, and the combined feature embeddings are passed to the classifier for classification.

\begin{figure}[H]
    \centering
    \includegraphics[width=0.4\linewidth]{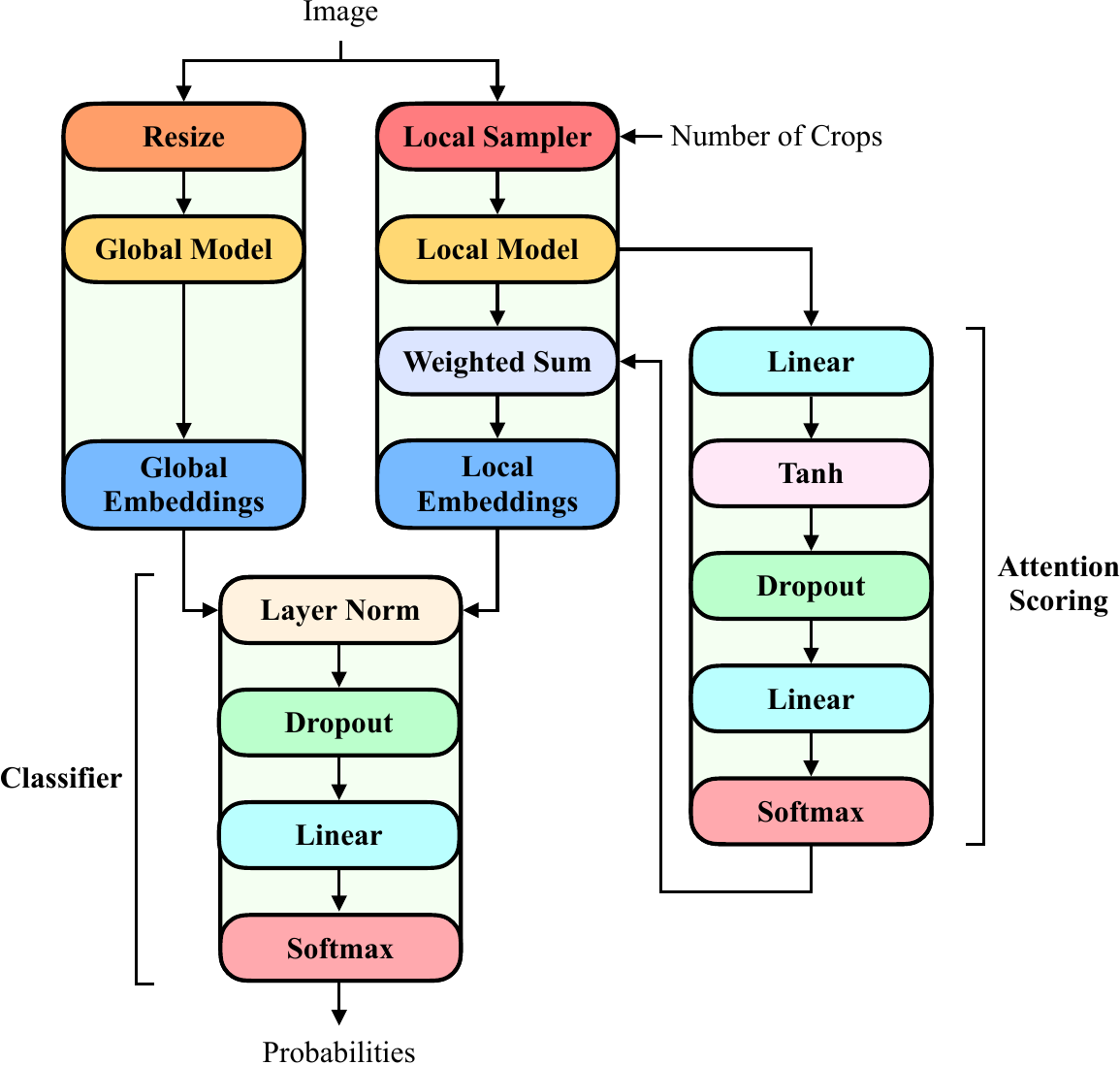}
    \caption{\textbf{GLASS architecture diagram.}}
    \label{fig:glass}
\end{figure}

\subsection{Multi-Scale Image Processing}
\subsubsection{Global Resize}
Given an RGB image $\boldsymbol{x}\in \mathbb{R}^{3\times H \times W}$ with resolution of $(H, W)$ pixels for natural $H, W \geq224$, we resize it to a square of $(224, 224)$ using bilinear interpolation, which is the input size required by the pre-trained models. 

\subsubsection{Local Crop Sampler}
    \begin{figure}[H]
    \centering
    \includegraphics[width=1\linewidth]{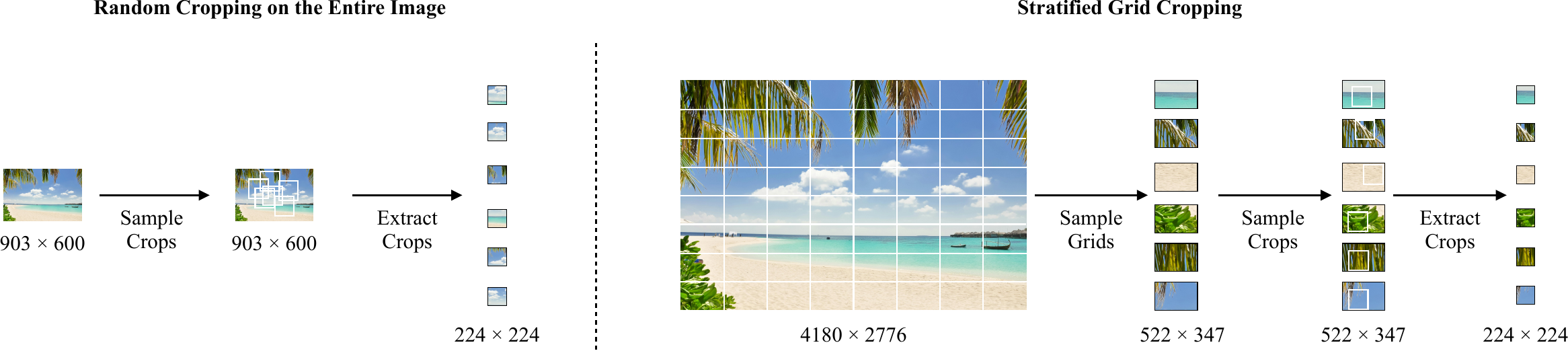}
    \caption{\textbf{Two cropping strategies of the local crop sampler.} Pixel dimensions are shown below each image.}
    \label{fig:local_crops_sampler}
\end{figure}

Given an RGB image $\boldsymbol{x}\in \mathbb{R}^{3\times H \times W}$ with resolution of $(H, W)$ pixels for natural $H, W \geq224$, and a number of crops $n\in \mathbb{N}$, we define a grid size of 

{\small
\[G = \min\left(\left\lfloor\frac{\min(H, W)}{224}\right\rfloor, 8\right) \text{ ,}\]}

partitioning the image into $G^2$ cells. The constant $8$ serves as an upper limit to the number of grid cells. This ensures computational efficiency and introduces additional spatial stochasticity through a larger cell size. Each cell has a size of $(\lfloor\frac{H}{G}\rfloor, \lfloor\frac{W}{G}\rfloor)$, ensuring that both dimensions are at least $224$ pixels. The local sampler has two cropping strategies for two different cases. When the number of grid cells is smaller than the number of crops, i.e. $G^2<n$, we sample $n$ crops uniformly at random over the entire image without the grid. When the number of grid cells is greater than or equal to the number of crops, i.e. $G^2\geq n$, we randomly select $n$ distinct grid cells and randomly sample one crop from each cell. All resulting crops from the two strategies are RGB images with a resolution of $(224,224)$, corresponding to a $224$-by-$224$ pixel square region of the original image at original resolution.

\begin{table}[htbp]
\centering
\footnotesize
\caption{\textbf{Expected percentage of covered area for various image sizes and number of crops $n$.}}
\label{tab:coverage}
    \begin{tabular}{lcccccccc}
    \toprule
    \textbf{Image Size} & \textbf{$n=2$} & \textbf{$n=4$} & \textbf{$n=6$} & \textbf{$n=8$} & \textbf{$n=10$} & \textbf{$n=12$} & \textbf{$n=14$} & \textbf{$n=16$} \\
    \midrule 
    224 $\times$ 224 & \entire{100.0} & \entire{100.0} & \entire{100.0} & \entire{100.0} & \entire{100.0} & \entire{100.0} & \entire{100.0} & \entire{100.0} \\
256 $\times$ 256 & \entire{83.9} & \entire{90.0} & \entire{92.8} & \entire{94.4} & \entire{95.4} & \entire{96.2} & \entire{96.7} & \entire{97.1} \\
640 $\times$ 480 & \grid{32.7} & \grid{65.3} & \entire{53.8} & \entire{60.5} & \entire{65.3} & \entire{69.0} & \entire{71.9} & \entire{74.2} \\
1024 $\times$ 768 & \grid{12.8} & \grid{25.5} & \grid{38.3} & \grid{51.0} & \entire{43.8} & \entire{48.8} & \entire{53.1} & \entire{56.8} \\
1280 $\times$ 720 & \grid{10.9} & \grid{21.8} & \grid{32.7} & \grid{43.6} & \entire{39.5} & \entire{44.5} & \entire{48.8} & \entire{52.6} \\
1920 $\times$ 1080 & \grid{4.8} & \grid{9.7} & \grid{14.5} & \grid{19.4} & \grid{24.2} & \grid{29.0} & \grid{33.9} & \grid{38.7} \\
2560 $\times$ 1440 & \grid{2.7} & \grid{5.4} & \grid{8.2} & \grid{10.9} & \grid{13.6} & \grid{16.3} & \grid{19.1} & \grid{21.8} \\
3840 $\times$ 2160 & \grid{1.2} & \grid{2.4} & \grid{3.6} & \grid{4.8} & \grid{6.0} & \grid{7.3} & \grid{8.5} & \grid{9.7} \\
5120 $\times$ 2880 & \grid{0.7} & \grid{1.4} & \grid{2.0} & \grid{2.7} & \grid{3.4} & \grid{4.1} & \grid{4.8} & \grid{5.4} \\
7680 $\times$ 4320 & \grid{0.3} & \grid{0.6} & \grid{0.9} & \grid{1.2} & \grid{1.5} & \grid{1.8} & \grid{2.1} & \grid{2.4} \\
    \bottomrule
    \end{tabular}
    
\smallskip
{\textit{Legend:} \gridOutside{Red cells} = stratified grid crop, \entireOutside{Blue cells} = random cropping on the entire image.}
\vspace{-5mm}
\end{table}

As shown in Table~\ref{tab:coverage}, low-resolution images typically use random cropping on the entire image, while for high-resolution images, the stratified grid cropping strategy is mainly used. 

\subsection{Backbone Models}
While capable of training from scratch, the GLASS architecture uses pre-trained models for efficiency and better performance. This paper uses three common backbone models for transfer learning, including ViT-Base/16~\citep{dosovitskiy2021imageworth16x16words}, ResNet-50~\citep{he2015deepresiduallearningimage}, and ConvNeXt-Tiny~\citep{liu2022convnet2020s}. The purpose of this paper is not to identify the optimal backbone model, but to demonstrate the effectiveness of the GLASS architecture across different types of backbone models.

\subsection{Attention Mechanism}
GLASS uses a trainable additive attention mechanism~\citep{bahdanau2016neuralmachinetranslationjointly} to assign importance scores to the sampled local crops. It is a two-layer fully connected multilayer perceptron with a Tanh~\citep{ilse2018attentionbaseddeepmultipleinstance} activation function and dropout~\citep{JMLR:v15:srivastava14a} for regularization. The attention layer processes the raw embedding outputs of each local crop from the backbone model into an importance score. Crops with a higher importance score indicate that they contain more valuable features for making a classification decision. All local crops are aggregated into a single representation, weighted by their importance score. This ensures that crops with a higher score contribute more to the classification. The parameters of the attention layer are learned through training to maximize its ability to identify important features.

\subsection{Classifier}
To obtain the final prediction, a lightweight classifier is attached to take both global and local feature embeddings as input and output probabilities. Global embeddings are the output of the global model, while the local embeddings are the attention-aggregated result from the local model. Since both backbone models already produce high-level feature embeddings, using a lightweight classifier can reduce unnecessary complexity and improve model generalization.

\subsection{Dataset}
We use a dataset of $12{,}000$ images, a subset of the ``ai-generated-images-vs-real-images''~\citep{ai-generated-images-vs-real-images_dataset} dataset. The dataset is evenly split between real images and AI-generated images. Three-quarters of the real images come from Pexels and Unsplash, while the remaining one-quarter comes from WikiArt. The AI-generated images are sourced from Stable Diffusion, MidJourney, and DALL-E, each contributing one-third to the fake class.

\begin{table}[H]
    \centering
    \footnotesize
    \caption{\textbf{Full Dataset Statistics.} The mean image size approximates a square aspect ratio due to the inclusion of both landscape and portrait images in the dataset.}
    \label{tab:datasets_stats}
    \begin{tabular}{lccc}
    \toprule
    \textbf{Class} & \textbf{Samples} & \textbf{Mean Height (pixel)} & \textbf{Mean Width (pixel)} \\
    \midrule
    real & 6,000 & 2,205 & 2,212 \\
    fake & 6,000 & 1,188 & 1,234 \\
    \bottomrule
    \end{tabular}
\end{table}

The full dataset is randomly split into $8{,}400$ ($70\%$) images for training, $1{,}800$ ($15\%$) images for validation, and $1{,}800$ images ($15\%$) for testing. All models, including standard transfer learning and GLASS-based models, are trained, validated, and tested on the same data with identical preprocessing.

\subsection{Hyperparameter Tuning}
The hyperparameters are fine-tuned using the Optuna~\citep{optuna_2019} framework. Standard transfer learning models have been allocated $50$ trials. GLASS-based models have twice as many hyperparameters; therefore, they have been allocated $100$ trials. Each trial is limited to a maximum of $25$ epochs. To improve efficiency, a Hyperband pruner is used to stop underperforming trials early. The hyperparameter search space is listed in Table~\ref{tab:transfer_hyperparameters} and Table~\ref{tab:glass_hyperparamters} in the appendix. Each of the two backbone models in the GLASS architecture is assigned a different learning rate and weight decay value. This ensures that each component of the GLASS architecture is optimized for its function.

\section{Results}
\subsection{Number of Crops}
\label{subsec:number_of_crops}
After obtaining the sets of hyperparameters that yield the highest validation accuracy, we retrain the models using these configurations, while varying only the number of crops and keeping the batch size fixed at $32$ images across all models. Using a fixed batch size for all models ensures a fair comparison of computational efficiency, where the memory usage and training time are not confounded by inconsistencies in batch size.
    
\begin{figure}[htbp]
    \centering
    \includegraphics[width=1\linewidth]{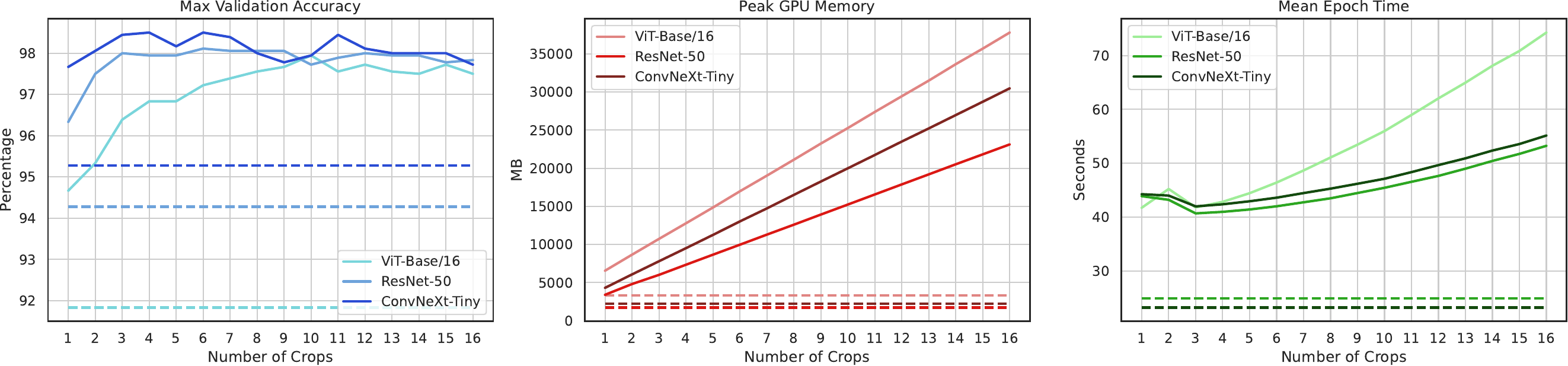}
    \vspace{-6mm}
    \caption{\textbf{Effect of the number of crops on validation accuracy, memory, and training time.} Solid lines represent results from GLASS-based models, while dotted line indicates results from standard transfer learning.}
    \label{fig:num_crops}
\end{figure}

The validation accuracy, memory usage, and training time remain constant for standard transfer learning models, since they only use the globally resized view without any local crops. The validation accuracy for GLASS-based models improves as the number of crops increases, eventually plateauing and then fluctuating after reaching a threshold. The initial increase in the number of crops could provide more spatial information to capture local details, while further increases introduce diminishing returns. This suggests that a moderate number of crops provides sufficient coverage of local features necessary to identify fine-grained artifacts and synthetic patterns. Both memory usage and training time increase progressively as the number of crops increases. 

\begin{table}[H]
\centering
\footnotesize
\caption{\textbf{Linear scaling of memory usage with number of crops $n$ from 1 to 16.} Peak GPU memory usage in MB is measured, and approximate linear regression fits with corresponding $R^2$ values are provided for GLASS models.}
\label{tab:num_crops_memory}
    \begin{tabular}{lccc}
    \toprule
    \textbf{Backbone Model} & \textbf{Linear Regression} & \textbf{$R^2$} & \textbf{Standard Transfer Learning} \\
    \midrule
    ViT-Base/16 & $4463.16 + 2081.92 \times n$ & $1.000$ & $3302.07$ \\
    \midrule
    ConvNeXt-Tiny & $2545.78 + 1744.40 \times n$ & $1.000$ & $2212.53$ \\
    \midrule
    ResNet-50 & $2088.69 + 1314.73 \times n$ & $1.000$ & $1722.31$ \\
    \bottomrule
    \end{tabular}
\end{table}

A linear composition of peak GPU memory is performed to analyze the effect of increasing numbers of crops, with a fixed batch size of 32 images. The memory usage for each number of crops is measured by the maximum peak GPU memory across all $25$ epochs. The intercept of the approximate linear regression fit represents the static memory component, while the slope indicates the marginal increase in memory per crop. The linear regression of all three models showed an $R^2$ value of $1.000$, indicating a perfect linear relationship between the peak GPU memory and the number of crops.

\begin{table}[H]
\centering
\footnotesize
\caption{\textbf{Linear scaling of training time with number of crops $n$ from 1 to 16.} The mean time per epoch in seconds is measured, and approximate linear regression fits with corresponding $R^2$ values are provided for GLASS models.}
\label{tab:num_crops_time}
    \begin{tabular}{lccc}
    \toprule
    \textbf{Backbone Model} & \textbf{Linear Regression} & \textbf{$R^2$} & \textbf{Standard Transfer Learning} \\
    \midrule
    ViT-Base/16 & $35.44 + 2.23 \times n$ & $0.948$ & $23.31$ \\
    \midrule
    ConvNeXt-Tiny & $40.05 + 0.82 \times n$ & $0.870$ & $23.23$ \\
    \midrule
    ResNet-50 & $39.12 + 0.74 \times n$ & $0.804$ & $24.91$ \\
    \bottomrule
    \end{tabular}
\end{table}

A similar linear decomposition is also performed for training time, measured by the mean time per epoch in seconds across all $25$ epochs. The intercept of the approximate linear regression fit represents the static epoch time component, while the slope indicates the marginal increase in epoch time per crop. The linear regression of the three models showed a mean $R^2$ value of $0.874$, indicating a strong linear relationship between the mean epoch time and the number of crops.

\subsection{Quantitative Performance of Final Models}
The training and validation datasets are combined and reshuffled for final model training, a total of $10{,}200$ images ($85\%$ of the full dataset). The final models are trained using their respective best-performing hyperparameters. GLASS-based models use their original optimal batch size obtained during hyperparameter tuning and the number of crops that yielded the highest validation accuracy in Subsection~\ref{subsec:number_of_crops}. These final hyperparameter configurations are reported in Table~\ref{tab:best_transfer_hyperparameters} and Table~\ref{tab:best_glass_hyperparameters} in the appendix.

\begin{table}[H]
\centering
\fontsize{8.7pt}{11pt}\selectfont
\caption{\textbf{Performance comparison of baseline models and GLASS models on the test dataset}. Results are reported as mean $\pm$ standard deviation over 5 runs with different random seeds, trained using the best hyperparameters. ECE refers to Expected Calibration Error. The best performances are bolded.}
\label{tab:main_results}
    \begin{tabular}{lcccccc}
    \toprule
    \textbf{Model} & \textbf{Accuracy (\%)} & \textbf{AUC} & \textbf{F1} & \textbf{Precision} & \textbf{Recall} & \textbf{ECE} \\
    \midrule
    Standard ViT-Base/16 & 93.244 $\pm$ 0.499 & 0.972 $\pm$ 0.006 & 0.932 $\pm$ 0.005 & 0.933 $\pm$ 0.005 & 0.932 $\pm$ 0.005 & \textbf{0.033 $\pm$ 0.003} \\
    \textbf{GLASS ViT-Base/16} & \textbf{97.533 $\pm$ 0.203} & \textbf{0.997 $\pm$ 0.000} & \textbf{0.975 $\pm$ 0.002} & \textbf{0.975 $\pm$ 0.002} & \textbf{0.975 $\pm$ 0.002} & 0.058 $\pm$ 0.003\\
    \midrule
    Standard ConvNeXt-Tiny & 96.133 $\pm$ 0.145 & 0.990 $\pm$ 0.002 & 0.961 $\pm$ 0.001 & 0.961 $\pm$ 0.001 & 0.961 $\pm$ 0.001 & \textbf{0.041 $\pm$ 0.002} \\
    \textbf{GLASS ConvNeXt-Tiny} & \textbf{98.611 $\pm$ 0.232} & \textbf{0.998 $\pm$ 0.001} & \textbf{0.986 $\pm$ 0.002} & \textbf{0.986 $\pm$ 0.002} & \textbf{0.986 $\pm$ 0.002} & 0.048 $\pm$ 0.001 \\
    \midrule
    Standard ResNet-50 & 93.656 $\pm$ 0.674 & 0.981 $\pm$ 0.002 & 0.937 $\pm$ 0.007 & 0.937 $\pm$ 0.007 & 0.937 $\pm$ 0.007 & \textbf{0.040 $\pm$ 0.005}\\
    \textbf{GLASS ResNet-50} & \textbf{97.656 $\pm$ 0.499} & \textbf{0.997 $\pm$ 0.001} & \textbf{0.977 $\pm$ 0.005} & \textbf{0.977 $\pm$ 0.005} & \textbf{0.977 $\pm$ 0.005} & 0.052 $\pm$ 0.004\\
    \bottomrule
    \end{tabular}
\end{table}

To ensure a fair comparison, GLASS-based models were trained for the same number of epochs as the standard models. Moreover, despite a search space containing double the number of hyperparameters, the computational budget was only twice as large for GLASS-based models, whereas, in practice, search complexity typically grows significantly faster than linearly. Nevertheless, GLASS-based models consistently achieved higher accuracy, AUC, F1 score, precision, and recall than standard transfer learning that uses only the globally resized view. However, they also consistently resulted in a higher Expected Calibration Error. 

\subsection{Weight Distribution}
\begin{figure}[htbp]
    \centering
    \includegraphics[width=1\linewidth]{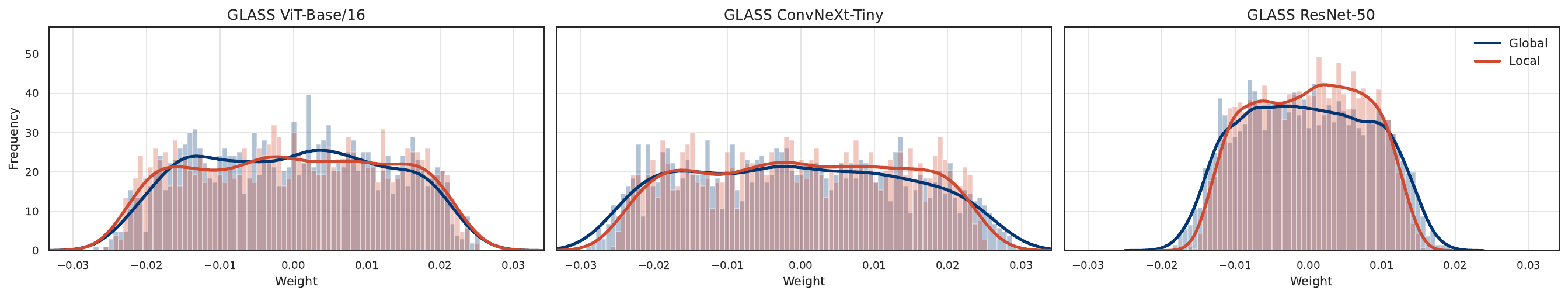}
    \caption{\textbf{Weight distributions in the linear layer of the classifier in the final GLASS-based models.} Solid lines are Kernel Density Estimation curves. The GLASS-based ViT-Base/16 and ConvNeXt-Tiny models each contain $3{,}072$ weights in their final linear layer, while the GLASS ResNet-50 model contains $8{,}192$ weights.}
    \label{fig:weight_distribution}
\end{figure}

The classifier in the GLASS architecture consists of a fully connected layer that takes global and local feature embeddings from the respective models to output two probabilities (real and AI-generated). Figure~\ref{fig:weight_distribution} illustrates how each model balances the contribution of global and local feature embeddings in this linear layer. Both global and local features are assigned similar weights, with similar distribution patterns; therefore, they are comparably relevant and informative for classification. Interestingly, the weight distribution range of the GLASS ResNet-50 model appears narrow, potentially due to a lower learning rate compared to the other models (hyperparameter values in Table~\ref{tab:best_glass_hyperparameters}).

\section{Conclusion}
The glass architecture introduced in this paper achieved an average increase of $3.59\%$ in accuracy compared to standard transfer learning that uses only the globally resized view. The results demonstrate that combining global and local features enhances the model's ability to capture local details and improves its predictive performance. Furthermore, the local sampler and the attention mechanism efficiently leverage local information without a significant increase in computational cost. Overall, these results suggest that GLASS is a promising architecture for AI-generated image detection, and potentially other tasks involving multi-scale feature integration.

\medskip
There are several limitations to this paper. Firstly, due to limited computational resources, the dataset diversity, training scale, and hyperparameter tuning were significantly constrained. A more extensive evaluation is necessary to fully demonstrate GLASS's performance. Secondly, the dataset used is from 2023 and may not fully reflect the capabilities of more recent generative models. Finally, a high classification accuracy was achieved even with standard transfer learning using only the resized image. This could suggest that the task may be relatively easy under the current experimental setting.

\nocite{NEURIPS2019_9015}
\bibliography{references}

\newpage
\appendix
\section{Appendix}
\subsection{Expected Sample Area of Local Crops}
Given an image with resolution $(H, W)$, where $H, W \geq p$, and crop size $(p, p)$ for $p\in \mathbb{N}$, we define a grid size of 
{\small
\[G = \min\left(\left\lfloor\frac{\min(H, W)}{224}\right\rfloor, 8\right) \text{ .}\]}

Next, we consider the following cases for the number of crops $n\in\mathbb{N}$.

\medskip
\textbf{Case 1:} $G^2 < n$

Each crop has an area of $p^2$, and overlaps are possible. The expected total covered area is

{\small
\[E[\text{covered area}] = \sum_{y=0}^{H-1} \sum_{x=0}^{W-1} \Big[1 - (1-p_{x, y})^n\Big] \approx WH\Big[1 - \Big(1-\frac{p^2}{WH}\Big)^n\Big] \text{ ,} \]}

where $p_{x, y}$ denotes the probability that pixel $(x, y)$ is included in a single crop:

{\small
\[p_{x, y} = \frac{a_y}{H-224+1} \times \frac{b_x}{W-224+1} \text{ ,}
\]}

with $a_y$ and $b_x$ represent the number of crops that cover row $y$ and column $x$, respectively:
{\footnotesize
\[
a_y = 
\begin{cases}
  y+1 & 0 \leq y \leq p-1\\
  p   & p \leq y \leq H-p\\
  H-y & H-p+1 \leq y \leq H-1
\end{cases} \text{ and }
b_x = 
\begin{cases}
  x+1 & 0 \leq x \leq p-1 \\
  p & p \leq x \leq W-p \\
  W-x & W-p+1 \leq x \leq W-1 \text{ .}
\end{cases}\]}

The expected percentage of the image covered is
{\small
\[E\text{[percentage covered]} = \frac{100}{WH}\sum_{y=0}^{H-1} \sum_{x=0}^{W-1} \Big[1 - (1-p_{x, y})^n\Big] \approx 100\times\Big[1 - \Big(1-\frac{p^2}{WH}\Big)^n\Big]\text{ .}\]}

\textbf{Case 2:} $G^2 \geq n$

Since overlaps are not possible, the expected total area covered is

{\small
\[E[\text{covered area}] = np^2 \text{ ,}\]}

and the percentage of area covered is

{\small
\[E[\text{percentage covered}] = 100\times \frac{np^2}{HW} \text{ .}\]}

\subsection{Hyperparameter Search Space}
\begin{table}[htbp]
\centering
\footnotesize
\caption{\textbf{Standard transfer learning models hyperparameter search space.} Square brackets represent the range of continuous variables, while curly brackets indicate categorical variables. All continuous values are sampled in the log domain.}
    \begin{tabular}{ll}
    \toprule
    \textbf{Hyperparameter} & \textbf{Search Space} \\
    \midrule
    Learning Rate & $[10^{-6}, 10^{-4}]$ \\
    Weight Decay  & $[10^{-7}, 10^{-3}]$ \\
    Dropout Rate  & $\{0.0, 0.1, 0.2, 0.3\}$ \\
    Batch Size & $\{32, 64, 128\}$ \\
    \bottomrule
    \end{tabular}
\label{tab:transfer_hyperparameters}
\end{table}

\begin{table}[htbp]
\centering
\footnotesize
\caption{\textbf{GLASS models hyperparameter search space.} Square brackets represent the range of continuous variables, while curly brackets indicate categorical variables. All continuous values are sampled in the log domain.}
    \begin{tabular}{ll}
    \toprule
    \textbf{Hyperparameter} & \textbf{Search Space} \\
    \midrule
    Global Learning Rate & $[10^{-6}, 10^{-4}]$ \\
    Local Learning Rate & $[10^{-6}, 10^{-4}]$ \\
    Attention and Classifier Learning Rate & $[5\times10^{-6}, 10^{-3}]$ \\
    Global Weight Decay & $[10^{-7}, 10^{-3}]$ \\
    Local Weight Decay & $[10^{-7}, 10^{-3}]$ \\
    Dropout Rate & $\{0.0, 0.1, 0.2, 0.3\}$ \\
    Batch Size & $\{16, 32, 64\}$ \\
    Number of Crops & $\{2, 4, 6, 8, 10\}$ \\
    \bottomrule
    \end{tabular}
\label{tab:glass_hyperparamters}
\end{table}

\subsection{Hyperparameter Tuning Plots}
\begin{figure}[H]
    \centering
    \includegraphics[width=0.87\linewidth]{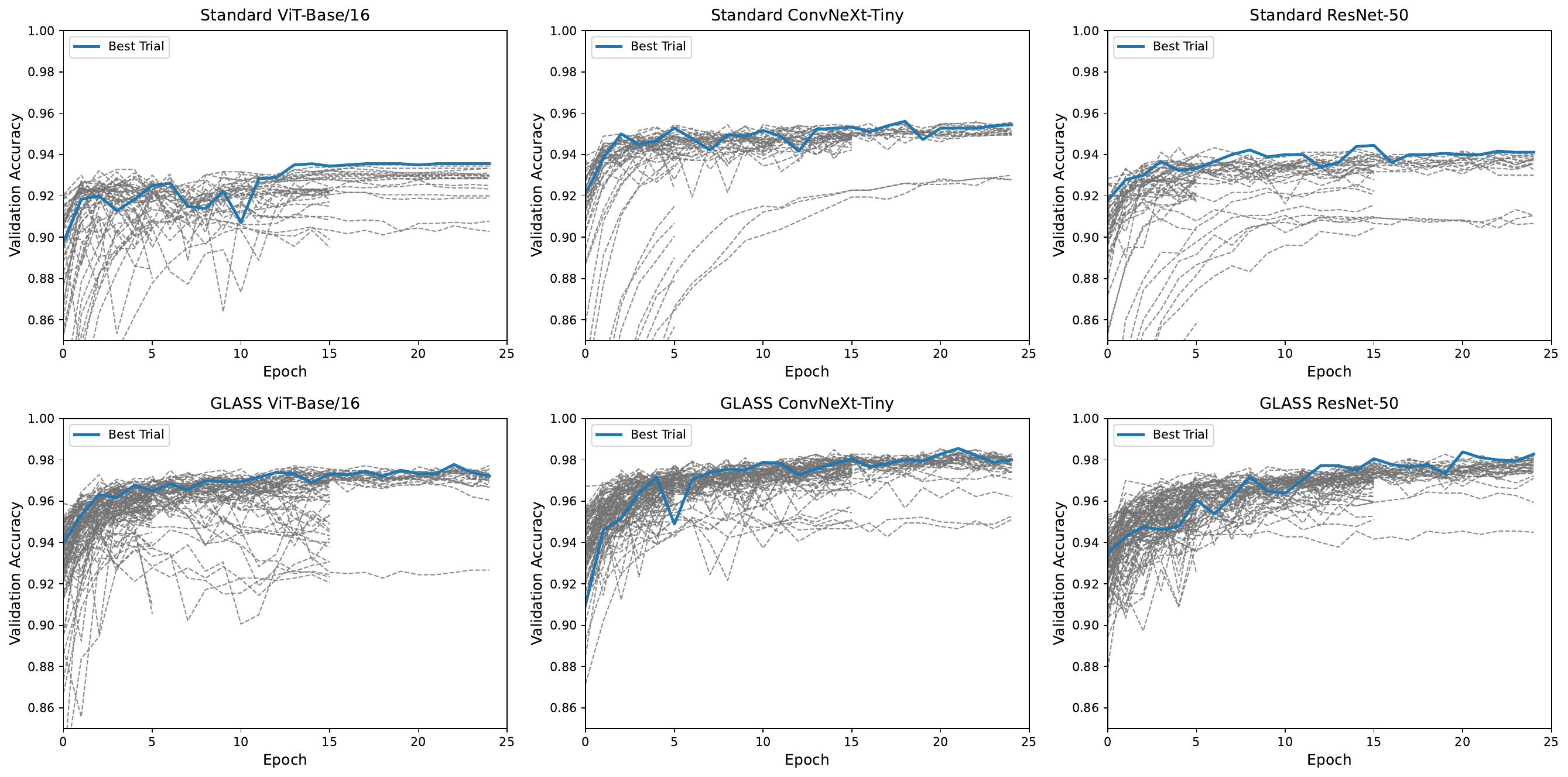}
    \caption{\textbf{Validation accuracy curves of all hyperparameter optimization trials for different models.}}
    \label{fig:hyperparameter_tuning_acc}
\end{figure}

\begin{figure}[H]
    \centering
    \includegraphics[width=0.87\linewidth]{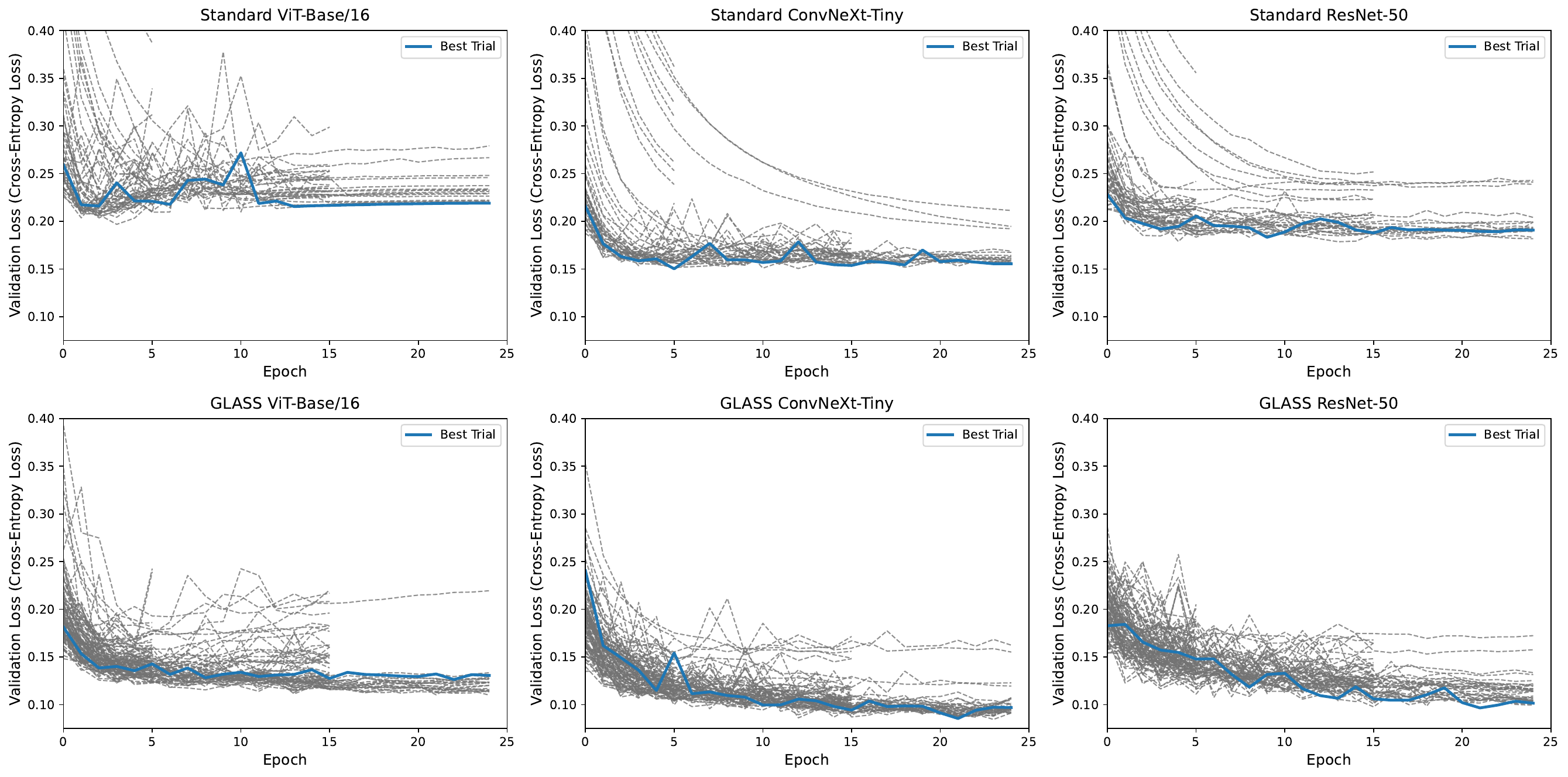}
    \caption{\textbf{Validation cross-entropy loss curves of all hyperparameter optimization trials for different models.}}
    \label{fig:hyperparameter_tuning_loss}
\end{figure}

\subsection{Best Hyperparameters}
\begin{table}[H]
\centering
\small
\caption{\textbf{Best hyperparameters for standard transfer learning models.} The displayed continuous values are rounded to three significant figures, while the final models were trained using full precision values.}
\begin{tabular}{llll}
\toprule
\textbf{Hyperparameter} & \textbf{Standard ViT-Base/16} & \textbf{Standard ConvNeXt-Tiny} & \textbf{Standard ResNet-50} \\
\midrule
Learning rate & $7.48 \times 10^{-5}$ & $7.12 \times 10^{-5}$ & $4.59 \times 10^{-5}$\\
Weight decay & $4.37 \times 10^{-4}$ & $2.13 \times 10^{-6}$ & $3.64 \times 10^{-7}$\\
Dropout rate & $0.1$ & $0.1$ & $0.1$\\
Batch size & $128$ & $128$ & $32$ \\
\bottomrule
\end{tabular}
\label{tab:best_transfer_hyperparameters}
\end{table}

\begin{table}[H]
\centering
\small
\caption{\textbf{Best hyperparameters for GLASS-based models.} The displayed continuous values are rounded to three significant figures, while the final models were trained using full precision values.}
\begin{tabular}{llll}
\toprule
\textbf{Hyperparameter} & \textbf{GLASS ViT-Base/16} & \textbf{GLASS ConvNeXt-Tiny} & \textbf{GLASS ResNet-50} \\
\midrule
Global learning rate & $1.58 \times 10^{-5}$ & $6.55 \times 10^{-6}$ & $2.16 \times 10^{-6}$\\
Local learning rate & $4.26 \times 10^{-5}$ & $1.00 \times 10^{-4}$ & $6.85 \times 10^{-5}$\\
Attention and classifier learning rate & $6.48 \times 10^{-5}$ & $1.68 \times 10^{-5}$ & $2.98 \times 10^{-5}$\\
Global weight decay & $3.18 \times 10^{-5}$ & $2.85 \times 10^{-6}$ & $1.79 \times 10^{-7}$\\
Local weight decay & $6.14 \times 10^{-6}$ & $3.88 \times 10^{-7}$ & $6.02 \times 10^{-7}$\\
Dropout rate & $0.3$ & $0.2$ & $0.3$\\
Batch size & $64$ & $32$ & $32$\\
Number of crops & $10$ & $4$ & $6$\\
\bottomrule
\end{tabular}
\label{tab:best_glass_hyperparameters}
\end{table}

\end{document}